\documentclass[journal]{IEEEtran}

\usepackage{float}
\usepackage{caption}
\usepackage{booktabs}
\usepackage{titlesec}

\newcommand{\setupSupplement}{
  \clearpage
  \setcounter{section}{0}
  \setcounter{figure}{0}
  \setcounter{table}{0}
  \renewcommand{\thesection}{\arabic{section}}
  \renewcommand{\thesubsection}{\arabic{section}.\arabic{subsection}}
  \renewcommand{\thefigure}{S\arabic{figure}}
  \renewcommand{\thetable}{S\arabic{table}}

  \titlespacing*{\section}{0pt}{1ex}{0.5ex}
  \titlespacing*{\subsection}{0pt}{0.5ex}{0.2ex}
}

\newcommand{\safeincludegraphics}[2][]{%
  \IfFileExists{#2}{\includegraphics[#1]{#2}}{%
    \fbox{\parbox[c][0.23\textheight][c]{0.9\linewidth}{\centering Missing figure file:\\ \texttt{\detokenize{#2}}}}%
  }%
}

\ifCLASSINFOpdf
  \usepackage[pdftex]{graphicx}
\else
\fi

\usepackage{dsfont}

\usepackage{amsmath}

\usepackage{array}
\usepackage{cite}
\usepackage{amssymb}

\title{A Nationwide Japanese Medical Claims Foundation Model: Balancing Model Scaling and Task-Specific Computational Efficiency}

\author{Nanae~Aratake\textsuperscript{1},
        Taisei~Tosaki\textsuperscript{1},
        Yuji~Okamoto\textsuperscript{1},
        Eiichiro~Uchino\textsuperscript{1},
        Masaki~Nakamura\textsuperscript{2},
        Nobutomo~Matsui\textsuperscript{3},
        Akiko~Hatakama\textsuperscript{4},
        and~Yasushi~Okuno\textsuperscript{1}%
\thanks{\textsuperscript{1}Graduate School of Medicine, Kyoto University, Kyoto, Japan.}%
\thanks{\textsuperscript{2}Medical Data Vision Co., Ltd., Tokyo, Japan.}%
\thanks{\textsuperscript{3}IQVIA Solutions Japan G.K., Tokyo, Japan.}%
\thanks{\textsuperscript{4}DeSC Healthcare, Inc., Tokyo, Japan.}%
\thanks{Corresponding author: Yasushi Okuno (e-mail: \texttt{okuno.yasushi.4c@kyoto-u.ac.jp}).}%
}

\usepackage{url}

\begin{document}
\maketitle

\begin{abstract}
Clinical risk prediction using longitudinal medical data supports individualized care. Self-supervised foundation models have emerged as a promising approach for leveraging large-scale unlabeled healthcare records. In natural language processing, scaling laws suggest that larger models achieve predictably lower pretraining losses, supporting the foundation model paradigm. However, for structured medical data---characterized by a limited vocabulary and sparse observations---whether increasing model size consistently improves downstream predictions is unclear, as most studies evaluate only a single model scale. In this study, we evaluated the relationship between model scale and downstream task performance for structured medical foundation models. Using a random sample (2.3 million patients, 32 hospitals) from a nationwide 519-hospital Japanese claims database, we pretrained encoder-only Transformers at five scales (2.2M--101M parameters) for disease incidence and medication prediction. Downstream performance saturated at task-dependent thresholds: disease prediction benefited from larger models (32M--101M), whereas medication prediction saturated at 11M, reducing pretraining time by 178 h. Across all tasks, the best-performing model consistently outperformed a Light Gradient Boosting Machine baseline in the area under the precision--recall curve. These findings indicate that, unlike the monotonically decreasing pretraining loss, the optimal model size varied depending on task characteristics. This task-dependent saturation provides practical guidance for balancing predictive performance and computational cost in structured medical foundation models.
\end{abstract}


\section{Introduction}
\IEEEPARstart{C}{linical} risk prediction, which estimates the likelihood of disease onset or therapeutic intervention, supports individualized care and resource allocation~\cite{epi_CVDrisk_assess_2018, epi_MultiCancerRisk_assess_2020}. Longitudinal medical data that capture evolving patient states have attracted growing interest for this purpose~\cite{Rajkomar_MultiEventPred_2018}.

Most existing clinical risk prediction models rely on supervised learning~\cite{PtExp_Selfsupervised_DisPred_DenoiseAutoEn_2016}. Yet a major bottleneck in supervised approaches is the curation of task-specific labels, which is costly~\cite{EHR_LabelMake_2016}. Acquiring sufficient labels is often infeasible for rare diseases or novel clinical endpoints. Self-supervised pretraining followed by fine-tuning with limited labeled data has emerged as a promising paradigm. This approach enables models to leverage large volumes of unlabeled clinical records before adapting to specific tasks.

In natural language processing (NLP), scaling laws demonstrate that pretraining loss decreases predictably with model size, data, and computation~\cite{kaplan_scaling_2020}. 
This predictability supports the foundation model paradigm~\cite{bommasani_foundation_2021}. 
However, the optimal model size that is most useful for downstream fine-tuning depends on the task characteristics and labeled data availability~\cite{chinchilla_taskScaling_2022}. 
In the medical domain, where labeled data tend to be scarce, scaling up pretrained models may not be equally beneficial for all downstream tasks. 
This concern is particularly relevant to structured medical data, which, unlike free text, possess a relatively limited vocabulary and sparse, heterogeneous observations. Such properties may render representation learning prone to early saturation, suggesting that scaling gains may not translate uniformly into downstream improvements.

Existing structured medical foundation models~\cite{BEHRT_2020, MedBERT_2021, CEHR_BERT_2021, Hi_BEHRT_2022} typically evaluate downstream performance using only one model, which is often the largest variant. 
As a result, the performance--cost tradeoff across model sizes remains unexamined. 
Furthermore, clinical tasks differ fundamentally in nature. 
Disease onset prediction may require longer-range contextual modeling of partially unobserved biological processes, whereas medication initiation may depend more heavily on recent information governed by clinical guidelines and prescription patterns. Thus, the model size most useful for downstream fine-tuning is likely to be task dependent.

Beyond model size selection, the choice of modeling approach itself requires justification. Transformers~\cite{attn_all_u_need_2017} have become the predominant architecture for medical foundation models, owing to their capacity for long-range sequence modeling via self-attention~\cite{BEHRT_2020, MedBERT_2021}. By contrast, gradient-boosted decision trees (GBDTs) remain strong baselines for structured data~\cite{Grinsztajn_TreeVsDL_2022}, and the practical value of pretrained Transformers must be judged by whether their gains justify the added computation. Therefore, a systematic evaluation that jointly considers the model size, task characteristics, and computational constraints is required.

In this study, we systematically evaluated the relationship between the model size and downstream performance of structured medical foundation models. We used a nationwide hospital-based claims/Diagnosis Procedure Combination (DPC) database from Medical Data Vision Co., Ltd. (MDV)~\cite{MDV_JMDC_2022, claims_2019}. This database was selected because its comprehensive 519-hospital coverage provides the clinical diversity essential for foundation models. From a random sample of approximately 2.3 million patients across 32 hospitals, we constructed token sequences integrating diagnosis and medication codes, age in days, and sex.

Using these sequences, we pretrained the encoder-only Transformer models~\cite{attn_all_u_need_2017, BERT_2018} at five scales spanning 2.2M--101M parameters. This specific range was chosen to systematically evaluate the trade-offs between predictive performance and computational efficiency. We then fine-tuned these models to two fundamentally different task types: one-year incident disease prediction and one-year new medication initiation prediction. 
The performance was compared with that of a Light Gradient Boosting Machine (LGBM)~\cite{LightGBM_2017} baseline to assess the tradeoff among predictive performance, model capacity, and actual pretraining time.

The main contributions of this study are as follows:
\begin{itemize}
\item Systematic evaluation across five scales (2.2M--101M parameters) revealed a task-dependent capacity ceiling. Disease prediction benefited from larger models, whereas medication prediction saturated at 11M, achieving a 76\% reduction in pretraining time (53.9 h vs. 232.2 h) without compromising predictive performance.

\item For each task, the task-optimal pretrained model consistently surpassed the LGBM baseline in area under the precision--recall curve (AUPRC), demonstrating that appropriately sized---not uniformly massive---foundation models justify their pretraining cost. Evaluating task-specific capacity ceilings is therefore essential for resource-efficient model development.

\end{itemize}

\section{Related Work}
\subsection{Scaling Trends in Medical Data}
Scaling research on medical foundation models has advanced primarily in clinical texts. GatorTron~\cite{GatorTron_2022}, NYUTron~\cite{NYUTron_2023} and Med-PaLM~\cite{MedPaLM_2023} show monotonic downstream improvements with model size, but these findings rely on the vast vocabulary and complex context of free text. Whether the same holds true for structured data with a limited vocabulary and few observations remains unclear. This study addresses this gap by examining the relationship between model size and downstream task performance, specifically for structured medical data.

\subsection{Structured Medical Foundation Models and Task Suitability}
BEHRT~\cite{BEHRT_2020} and Med-BERT~\cite{MedBERT_2021} demonstrated the value of structured medical foundation models, but evaluated only a single model size, leaving the saturation point of downstream performance relative to computing unexplored. TransformEHR~\cite{TransformEHR_2023} proposes an encoder--decoder generative framework for disease outcome prediction. GenHPF~\cite{GenHPF_2023} introduced a multitask multisource predictive framework. However, both studies focused on architectural design rather than the relationship between model scale and downstream performance. EHRSHOT~\cite{EHRSHOT_2024} examined scaling with respect to data volume and prediction horizons in generative settings. Recently, Waxler et al.~\cite{curiosity_2025} compared multiple model sizes for generative medical event predictions in zero-shot settings. Although recent studies have explored adapting pretrained language models for tabular prediction~\cite{TP-BERTa_2024}, a systematic analysis of model size versus downstream performance for encoder-only architectures with fine-tuning of medical data remains lacking. The present study fills this gap by comparing disease prediction tasks, which require complex contextual understanding, with medication prediction tasks, which exhibit stronger regularity. This comparison provides insights into task-appropriate model size selection.

\subsection{Deep Learning versus GBDTs for Structured Data}
GBDTs frequently match or exceed the performance of deep learning in tabular data~\cite{Gorishniy_FT-Transformer_2021, Grinsztajn_TreeVsDL_2022}. For clinical deployment, the adoption of a large Transformer is justified only if the performance gain offsets the computational overhead. This study identified the model size threshold at which Transformers surpass GBDTs. These thresholds offer practical guidance for model size selection in clinical settings.

\section{Methods}
\begin{figure*}[t]
  \centering
  \includegraphics[width=\textwidth]{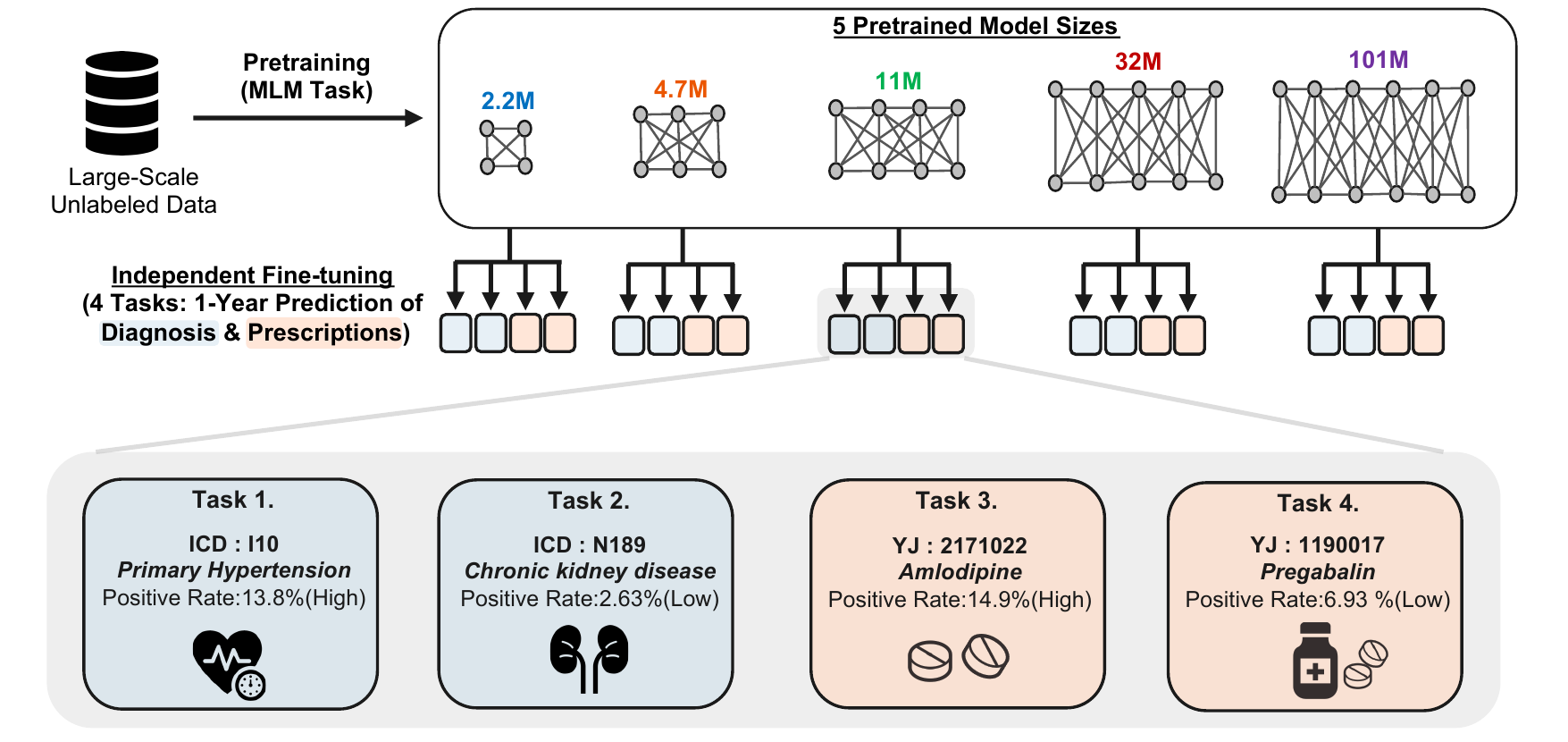}
  \caption{Overview of pretraining and fine-tuning. Masked language modeling (MLM) pretraining was performed on token sequences pairing clinical codes with age in days from approximately 2.3 million patients. Five encoder-only Transformers (2.2M--101M parameters) were independently fine-tuned on two disease prediction tasks (primary hypertension: 13.8\%; chronic kidney disease: 2.63\%) and two medication prediction tasks (amlodipine: 14.9\%; pregabalin: 6.93\%). Percentages in parentheses indicate the positive rate for each target event.}
  \label{fig:overview}
\end{figure*}

We pretrained encoder-only Transformers on patient-level sequences derived from nationwide claims/DPC data and fine-tuned them for incident disease prediction and new medication initiation prediction (Fig.~\ref{fig:overview}). 

Diagnoses were encoded using the International Statistical Classification 
of Diseases and Related Health Problems, 10th Revision (ICD-10), the 
standard diagnostic classification maintained by the World Health 
Organization. Medications were encoded using the YJ code, a 12-character 
individual drug code administered within the Japanese National Health 
Insurance drug pricing system by the Ministry of Health, Labour and 
Welfare (MHLW).

To consider the effect of label imbalance on predictive performance, we selected targets with different positive rates for both disease and medication prediction: primary hypertension (ICD-10: I10, 13.8\%), chronic kidney disease (ICD-10: N189, 2.63\%), amlodipine (YJ: 2171022, 14.9\%), and pregabalin (YJ: 1190017, 6.93\%). 
The performance was evaluated using the area under the receiver operating characteristic curve (AUROC) and AUPRC. These metrics were specifically chosen because AUROC serves as a standard measure of overall discriminative ability, whereas AUPRC is particularly crucial for accurately reflecting classification performance in clinical tasks characterized by low positive event rates.
 
\subsection{Dataset}
\subsubsection{Preprocessing}
We used anonymized outpatient and inpatient records from the MDV claims/DPC database. As introduced, we utilized a random sample of 32 facilities from the 519-facility parent database. This dataset includes diagnosis codes, medication codes, sex, and age in days (Table~\ref{tab:dataset}). To avoid potential training instability arising from extremely short sequences, we restricted the analysis to patients with at least two diagnoses or medication codes ($n = 2{,}294{,}687$). To improve the training stability and prioritize the learning of frequently occurring codes, we excluded diagnosis and medication codes occurring fewer than 10{,}000 times across the entire observation period. For YJ (medication) codes, only the first seven digits were used to collapse product-level differences to approximate pharmacological action levels. After these filtering steps, the vocabulary comprised 936 diagnoses and 994 medication codes. 

\subsubsection{Ethical Statement}
This was a retrospective observational study using anonymously processed data provided by MDV under a data use agreement. The Japanese Act on the Protection of Personal Information (enacted in 2003 and amended in 2022) defines anonymously processed data as information that have been irreversibly processed, rendering it technically and legally impossible to identify specific individuals. The restoration of such data to their original state is strictly prohibited by law. The acquired data were managed on an access-controlled university server restricted to authorized researchers.
The Ethical Guidelines for Life Sciences and Medical Research Involving Human Subjects in Japan (issued in 2021 and partially amended in 2023) exclude studies that solely use anonymously processed data from the scope of ethical review. In accordance with these national guidelines, the Ethics Committee of Kyoto University Graduate School and Faculty of Medicine, Kyoto University Hospital (http://www.ec.med.kyoto-u.ac.jp) maintains an institutional policy stating that studies relying exclusively on preexisting anonymously processed data do not require an application for ethical review, barring exceptional circumstances. Consequently, ethical approval for this study was formally waived based on this institutional policy, and the requirement to obtain informed consent from the study participants was not applicable.

\begin{table}[t]
\caption{Dataset Overview}
\label{tab:dataset}
\centering
\begin{tabular}{ll}
\hline
Data source & Medical Data Vision Co., Ltd. (MDV) \\
Number of hospitals & 32 (sampled from 519 in Japan) \\
Observation period & April 2014 -- November 2024 \\
Patients (total) & 2{,}445{,}160 \\
Patients (codes $\ge 2$) & 2{,}294{,}687 \\
Sex ratio & Male: 47.6\%, Female: 52.4\% \\
ICD-10 code coverage & 99.5\% \\
YJ medication code coverage & 85.4\% \\
Age and sex coverage & 100\% \\
\hline
\end{tabular}
\end{table}

\begin{figure*}[t]
  \centering
  \includegraphics[width=\linewidth]{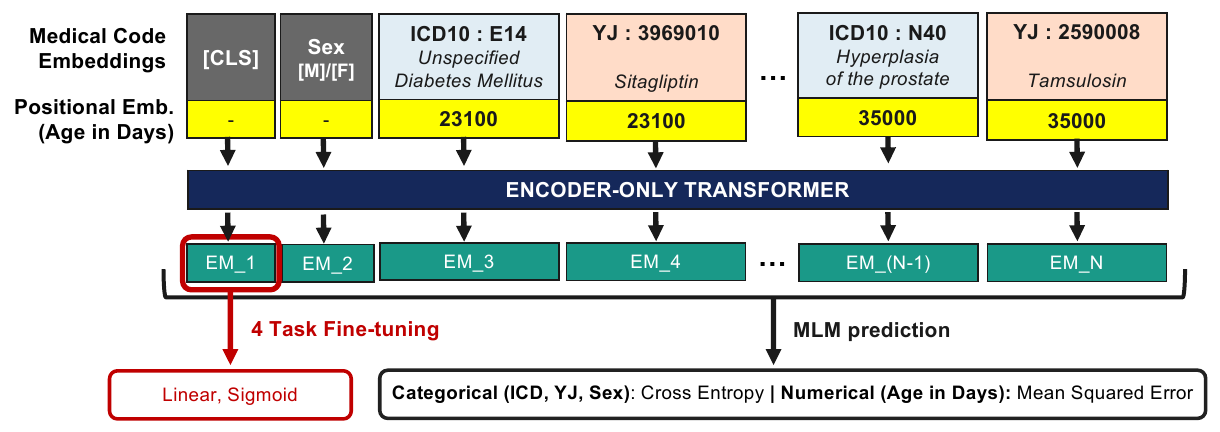}
  \caption{Input-sequence construction and training objectives. Each clinical event token pairs a code (ICD-10 or YJ) with age in days. The embedding is the element-wise sum of the code embedding and the age-in-days embedding. The sequence begins with [CLS] (for classification) and a sex token ([M]/[F]), followed by chronologically ordered event tokens. For pretraining, we employ masked language modeling (MLM), applying cross-entropy loss to categorical features (medical codes and sex) and mean squared error (MSE) to the numerical feature (age in days).}
  \label{fig:token}
\end{figure*}

\subsection{Token Sequence Construction}
Following previous work~\cite{BEHRT_2020, MedBERT_2021}, each patient's clinical history was represented by a single chronologically ordered sequence (Fig.~\ref{fig:token}). 
For patient $p$, the sequence $\mathbf{v}_p$ begins with a classification control token [CLS], followed by a sex attribute token ([M] or [F]) and chronologically ordered clinical tokens from position three onward. Each clinical token at position $i$ ($i \ge 3$) is defined as a pair $v^i_p = (z^i_p, a^i_p)$ comprising a categorical clinical code $z^i_p$ and numerical age in days $a^i_p$ at the time of recording. Denoting the sequence length for patient $p$ as $n$, the token sequence is
\begin{equation}
\mathbf{v}_p = (\text{[CLS]},\; \text{[M] or [F]},\; (z^3_p, a^3_p),\; \dots,\; (z^n_p, a^n_p)).
\end{equation}
The overall sequence statistics are summarized in the Supplementary Material.

\subsection{Input Embedding Construction}
The token sequence $\mathbf{v}_p$ defined in (1) is mapped to a $d$-dimensional vector sequence $\mathbf{E}_p = (\mathbf{E}^1_p, \mathbf{E}^2_p, \dots, \mathbf{E}^n_p) \in \mathbb{R}^{n \times d}$.
At each position $i$, the input embedding $\mathbf{E}_p^i$ comprises a categorical embedding corresponding to the discrete token identity and an embedding of age in days, which is a continuously valued attribute associated with each clinical event.

\paragraph{Categorical Embeddings}
We used separate embedding matrices for the special tokens, diagnosis codes, and medication codes. 
The token type at position $i$ is denoted $\mathbf{c}^i_p \in \{0, 1, 2\}$, where the type of vector for the entire sequence is $\mathbf{c}_p = (\mathbf{c}^1_p, \dots, \mathbf{c}^n_p) \in \{0, 1, 2\}^n$. Here, $\mathbf{c}^i_p = 0$ denotes special tokens, $\mathbf{c}^i_p = 1$ denotes diagnoses, and $\mathbf{c}^i_p = 2$ denotes medications. By letting $V_\text{l}$, $V_\text{icd}$, and $V_\text{yj}$ denote the respective vocabulary sizes, the corresponding embedding matrices are
\begin{equation}
\mathbf{W}_\text{l} \in \mathbb{R}^{V_\text{l} \times d},\quad \mathbf{W}_\text{icd} \in \mathbb{R}^{V_\text{icd} \times d},\quad \mathbf{W}_\text{yj} \in \mathbb{R}^{V_\text{yj} \times d}.
\end{equation}

\paragraph{Age-in-Days Embedding}
Age (in days) was used as the continuous temporal attribute for each clinical event. 
This scalar must be mapped into the same latent space as the categorical embeddings; hence, we encoded it using Piecewise Linear Encoding (PLE)~\cite{Gorishniy_PLE_2022}, which partitions the observed range into intervals and linearly interpolates within each interval.
This encoding can effectively capture the distributional properties of numerical features.

Let $a \in \mathbb{R}$ denote a pre-standardized continuous scalar. 
To define $\text{PLE} \colon \mathbb{R} \to \mathbb{R}^d$, we partitioned the value range into $d$ nonoverlapping intervals $B_t = [b_{t-1}, b_t)$ for $t = 1, \dots, d$, where $b_0 = \min(a)$ and $b_d = \max(a)$. 
The interval endpoints $\{b_t\}$ were determined from the training data using the quantiles of the empirical distribution:
\begin{equation}
b_t = Q_{t/d}(\{a^{(r)}\}),\quad t = 1, \dots, d-1,
\end{equation}
where $Q_{t/d}$ is the $t/d$-quantile of the training data.

Using these quantile endpoints, scalar $a$ was mapped to a $d$-dimensional embedding that reflected its statistical distribution. The $k$th component of $\text{PLE}(a)$ was
\begin{equation}
u_k(a) \triangleq \begin{cases}
0, & k > j \\
1, & k < j \\
\displaystyle\frac{a - b_{j-1}}{b_j - b_{j-1}}, & k = j
\end{cases}
\quad (k = 1, \dots, d),
\end{equation}
where $j$ satisfies $b_{j-1} \le a < b_j$. When combining PLE with other embeddings through summation, an affine transformation is recommended to align the representation spaces.

\paragraph{Final Input Embedding}
By applying PLE to $a^i_p$ with an affine transformation ($\mathbf{W}_{\text{age}} \in \mathbb{R}^{d \times d}$, $\mathbf{b} \in \mathbb{R}^d$), the embedding at position $i$ is
\begin{equation}
\begin{split}
\mathbf{E}^i_p = {} & \mathds{1}(\mathbf{c}^i_p\!=\!0)\, \mathbf{W}_{\text{l}}[z^i_p]
  + \mathds{1}(\mathbf{c}^i_p\!=\!1)\, \mathbf{W}_{\text{icd}}[z^i_p] \\
  & + \mathds{1}(\mathbf{c}^i_p\!=\!2)\, \mathbf{W}_{\text{yj}}[z^i_p]
  + \mathds{1}(\mathbf{c}^i_p\!\ge\!1)\, (\mathbf{W}_{\text{age}}\, \text{PLE}(a^i_p) + \mathbf{b}),
\end{split}
\end{equation}
where $\mathbf{W}[\cdot]$ denotes the embedding lookup and $\mathds{1}(\cdot)$ is the indicator function. The Transformer $f_{\theta}$ receives $\mathbf{E}_p = (\mathbf{E}^1_p, \dots, \mathbf{E}^n_p)$ and produces output $\mathbf{H}_p$.

\subsection{Training Procedure}
\paragraph{Pretraining}
We employed masked language modeling (MLM) with a maximum sequence length of 8{,}000. Sequences longer than this limit were truncated to retain the earliest 8{,}000 chronologically ordered tokens. Importantly, 99.97\% of the patient sequences in our dataset fell within this limit, ensuring minimal loss of clinical information. Fifteen percent of the tokens were randomly masked, and the model was trained to recover the true values.
As each token $v^i_p = (z^i_p, a^i_p)$ comprises a categorical numerical pair, the loss was constructed using distribution-appropriate components. 
The final loss is the unweighted sum of three independent terms: cross-entropy losses for diagnosis codes and medication codes (each with a vocabulary-specific output head and softmax) and mean squared error (MSE) for age in days. 
The model acquires general-purpose representations of each patient's medical context by jointly learning the categorical clinical codes and numerical ages.

\paragraph{Fine-tuning}
For each task, only the sequence preceding the target event date was used as the input. A binary classification was performed to predict whether the target diagnosis or medication would occur within one year. 
Operationally, the hidden state of the [CLS] token was extracted from the output of the pretrained model and passed through a single-layer output head with a sigmoid activation function to compute the event probability. 
Binary cross-entropy loss was used with full-parameter fine-tuning. To simulate limited-label conditions representative of real clinical settings, we varied the number of labeled patients between 100, 500, and 1{,}000.

\subsection{LGBM Baseline}
As a strong tabular baseline, LGBM was trained on count-based aggregated history features derived from the same input window as that of the transformer. 
Diagnosis and medication codes observed in the chronologically ordered history were converted into per-code counts, and sex and age in days at the index date were appended, yielding 1{,}932 features. 
This baseline preserves aggregate historical information but does not explicitly model the token order. 

\section{Experiments}
\subsection{Dataset Splitting}
The dataset was randomly split into pretraining (70\%), validation (10\%), test (10\%), and fine-tuning (10\%) subsets (seed 42), with no patient overlap. The pretraining partition comprised 1{,}606{,}268 patients, with 229{,}565 validation, 229{,}293 test, and 229{,}561 fine-tuning patients; each downstream subset exceeded 229{,}000 patients, providing sufficient statistical power for evaluation.

From the fine-tuning cohort (229{,}561), patients satisfying the following criteria were extracted per task: (i) observation period $\ge$1 year, (ii) positive cases with a one-year window before the first target event, and (iii) negative cases with no target event and a secured one-year window. Balanced samples (1:1) of 100, 500, and 1{,}000 labeled patients were drawn.
To examine the robustness, we generated three patient-level dataset realizations (seeds 42, 123, and 456). 
In each realization, class balancing (1:1 positive: negative) was applied only to the limited-label training subset of sizes 100, 500, or 1{,}000 fine-tuning dataset. 

\subsection{Computational Cost Evaluation}
We computed the theoretical floating-point operations (FLOPs) for each model and compared them against the test loss to evaluate the relationship between the model size and computational cost. Computations were performed on the Miyabi system at the University of Tokyo, utilizing 16 NVIDIA GH200 nodes for pretraining and a single node for fine-tuning. We also measured the actual pretraining wall-clock time (cumulative active training and validation time to reach the minimum validation loss, excluding queuing delays).

\subsection{Downstream Task Evaluation}
For each model size, we compared the downstream task performance of the pretrained model with that of a randomly initialized (from-scratch) model with identical architecture, both fine-tuned on the four downstream tasks. 
This comparison isolates the contribution of pretraining from the representational capacity of the architecture itself. Three independent patient-level datasets (seeds 42, 123, and 456) were used per label count condition; from-scratch models additionally used three initialization seeds (1001, 1002, and 1003) to reduce initialization dependence, with the mean performance reported.
The metrics AUROC and AUPRC were compared across all five pretrained models and LGBM for each label count condition. 

\subsection{Hyperparameter Settings}
\paragraph{Transformer Pretraining}
The hidden dimensions varied across 128, 256, 512, 1024, and 2048, with layers fixed at four, attention heads at 16, batch size at 512, and the maximum sequence length at 8{,}000. Default learning rates of 1e-4 and 3e-4 were initially applied to all models. While smaller models (128, 256, 512) trained stably with these defaults, the 1024 and 2048 models exhibited early overfitting; therefore, a grid search over \{5e-5, 3e-5, 1e-5, 5e-6\} was performed for these two larger models.

\paragraph{Transformer Fine-tuning}
Batch size: 8, learning rate: 1e-5, maximum epochs: 20 (for 100 patients) or 40 (for 500/1{,}000). Early stopping was triggered when the validation AUROC improvement $<$0.001 for two consecutive epochs.

\paragraph{LGBM}
Five-fold cross-validated grid search over learning rate: \{0.01, 0.05, 0.1\}, maximum depth: \{3, 4, 6\}, leaves: \{7, 15, 31\}, subsampling: \{0.8, 1.0\}, column subsampling: \{0.8, 1.0\}, and trees: \{50, 100, 200\}.

\section{Results}
\subsection{Pretraining Loss across Model Sizes}
\begin{figure}[t]
  \centering
  \includegraphics[width=\linewidth]{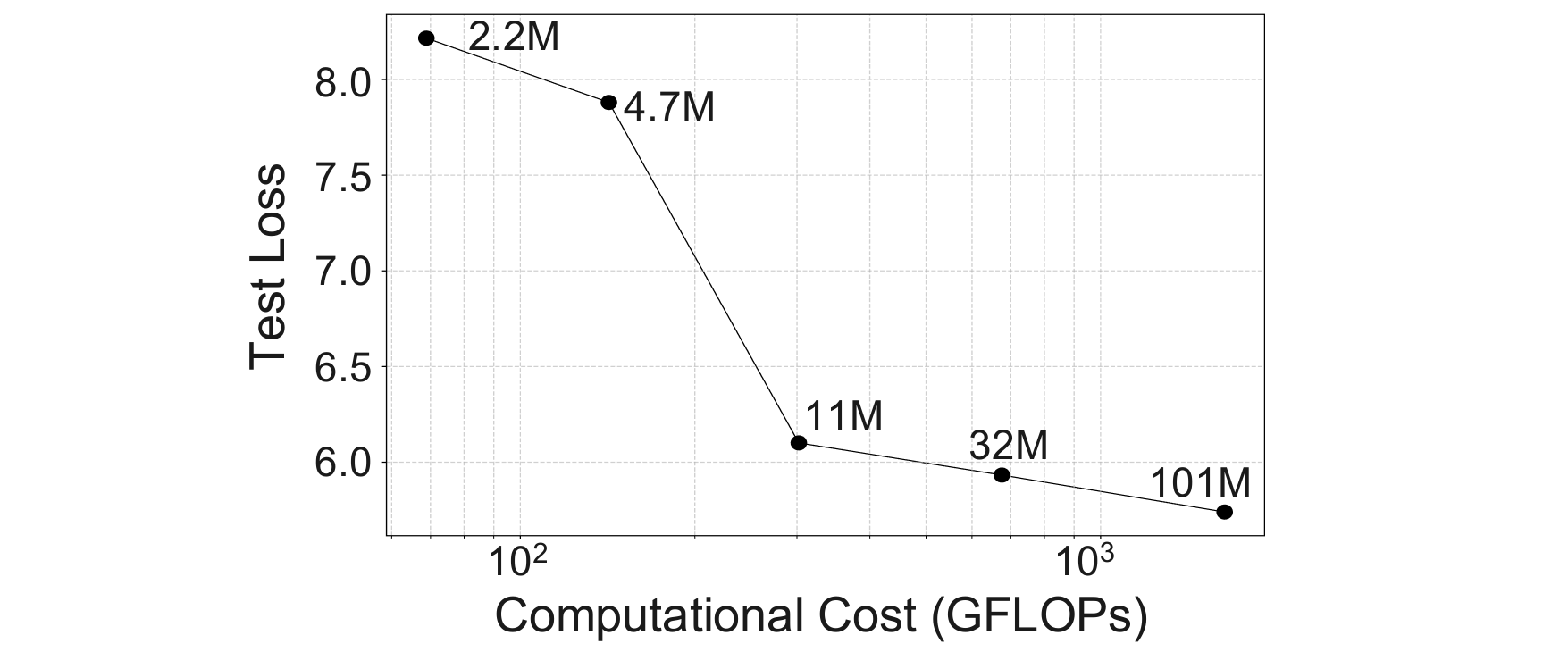}
  \caption{Pretraining loss versus computational cost. The test loss decreases with increasing GFLOPs as the model size grows. Each plot corresponds to a model size where $M = 10^6$ indicates the number of parameters. GFLOPs denotes $10^9$ floating-point operations.}
  \label{fig:scaling}
\end{figure}

Fig.~\ref{fig:scaling} and Table~\ref{tab:scale} show the relationship between the computational cost measured in giga floating-point operations (GFLOPs) and the test loss for the five transformer models pretrained at their respective optimal learning rates. GFLOPs represent the theoretical operations required for one forward/backward pass expressed in units of $10^9$ FLOPs. The computational cost increased exponentially with model size, whereas the test loss decreased substantially from the 11M model onward and continued to improve in the 101M model. Alongside the exponential increase in theoretical GFLOPs, the actual pretraining time increased substantially from 14.4 h for the 2.2M model to 232.2 h for the 101M model (Table \ref{tab:scale}).

\begin{table*}[t]
\caption{Model scales and computational costs for pretraining}
\label{tab:scale}
\centering
\begin{tabular}{cccccc}
\hline
Hidden Dimension $d$ & Parameters ($M$)$^{\mathrm{a}}$ & GFLOPs & Optimal Learning Rate & Pretraining Time (hours)$^{\mathrm{b}}$ & Test Loss \\
\hline
128  & 2.16   & 68.90   & 3e-4 & 14.4  & 8.22 \\
256  & 4.65   & 142.20  & 1e-4 & 25.0  & 7.88 \\
512  & 11.40  & 302.20  & 1e-4 & 53.9  & 6.10 \\
1024 & 31.97  & 675.70  & 3e-5 & 107.3 & 5.93 \\
2048 & 101.43 & 1636.52 & 1e-5 & 232.2 & \textbf{5.74} \\
\hline
\multicolumn{6}{l}{\footnotesize $^{\mathrm{a}}$ $M = 10^6$.} \\
\multicolumn{6}{l}{\footnotesize $^{\mathrm{b}}$ Wall-clock time required to reach the minimum validation loss.} \\
\end{tabular}
\end{table*}

\subsection{Fine-tuning Data Statistics}
Table~\ref{tab:fine_median_iqr} summarizes the input-sequence statistics for each task. Medication prediction tasks had longer sequences, more time points, and longer observation periods than disease prediction tasks.

\begin{table*}[t]
\caption{Input sequence statistics for four prediction tasks (Median [InterQuartile Range])}
\label{tab:fine_median_iqr}
\centering
\begin{tabular}{lcccc}
\hline
Statistic & Primary Hypertension & Chronic Kidney Disease & Amlodipine & Pregabalin \\
\hline
Sequence length (tokens) & 49 [17--142]  & 81 [25--277] & 92 [31--287] & 116 [37--364] \\
Time points (days) & 15 [6--37] & 22 [9--60] & 26 [10--63] & 31 [12--75] \\
Observation period (days) & 1431 [690--2539] & 1737 [889--2891] & 1766 [872--2916] & 1883 [989--3017] \\
\hline
\end{tabular}
\end{table*}

\begin{figure}[t]
  \centering
  \includegraphics[width=\linewidth]{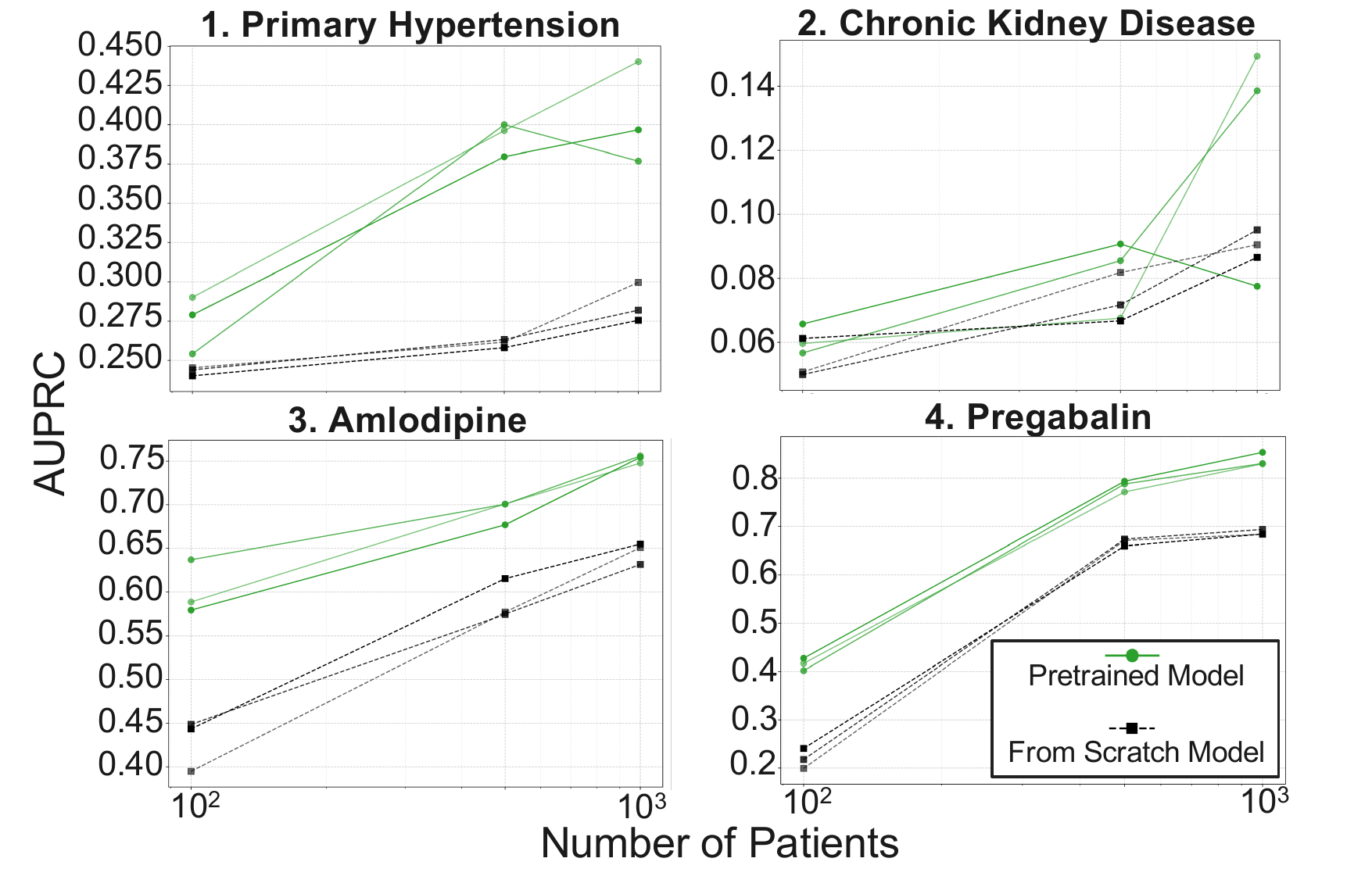}
  \caption{Pretrained vs.\ from-scratch AUPRC at the 11M scale across four tasks, with labeled patient counts of 100, 500, and 1{,}000. Three dataset splits shown per condition. The pretrained model (green) generally outperforms the from-scratch model (black). The x-axis is shared across all panels.}
  \label{fig:512dim_pretrained_vs_scratch}
\end{figure}

\begin{figure*}[t]
  \centering
  \includegraphics[width=\linewidth]{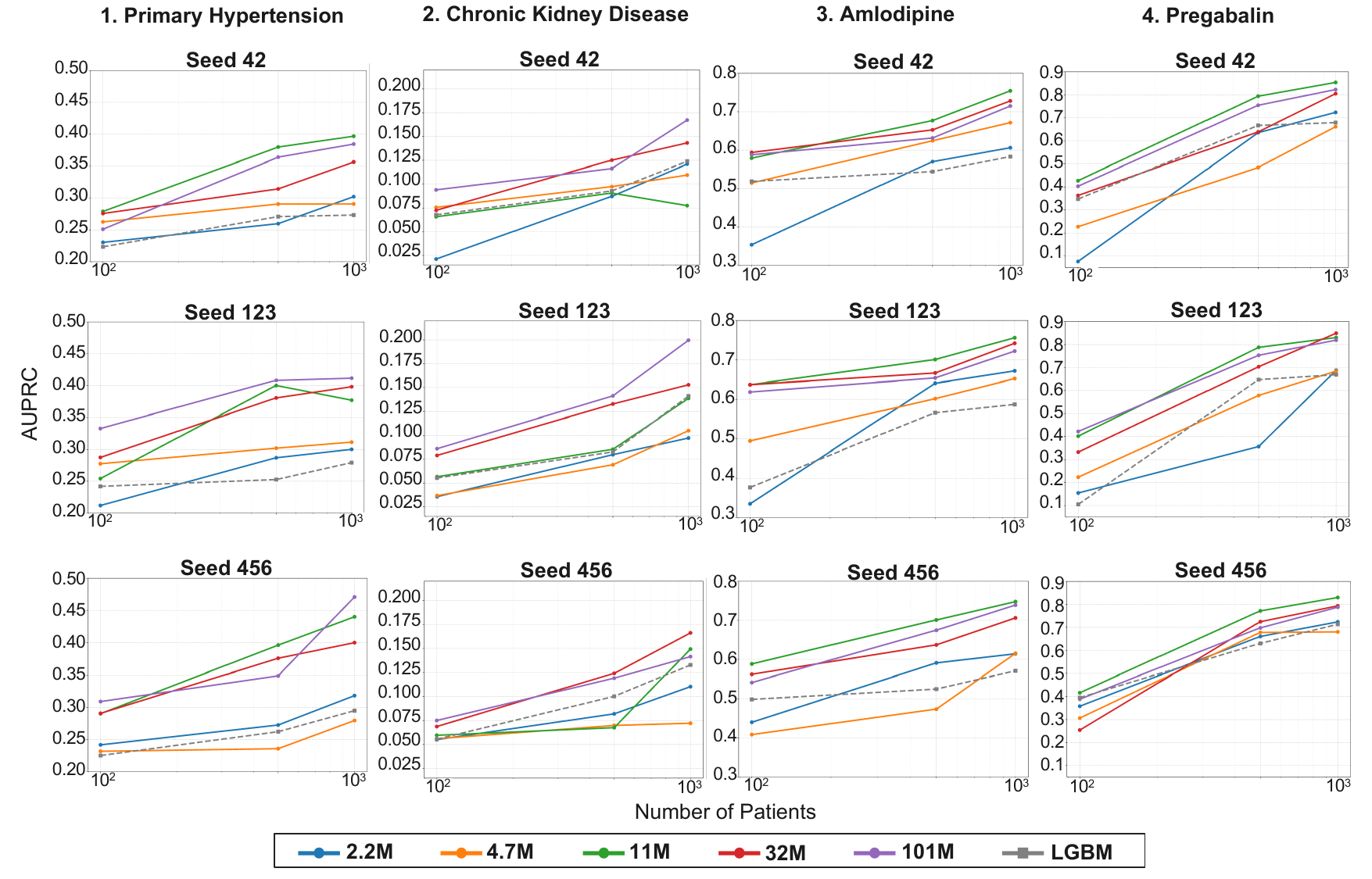}
  \caption{AUPRC of five pretrained Transformers and LGBM across four tasks. Disease prediction benefits from the 32M/101M model; medication prediction saturates at 11M. The task-optimal model consistently outperforms LGBM.}
  \label{fig:downstream}
\end{figure*}

\subsection{Pretrained versus From-scratch Models}
Fig.~\ref{fig:512dim_pretrained_vs_scratch} compares the pretrained and from-scratch AUPRC at the representative 11M scale; the results for all five sizes are shown in Supplementary Fig.~S2. Although the benefits of pretraining were occasionally insufficient for certain model sizes, label counts, or task types, the pretrained models generally outperformed their from-scratch counterparts. Across all tasks and model sizes, the performance improved consistently with increasing labeled data. Medication prediction tasks tended to achieve a higher overall performance than disease prediction tasks.

\subsection{Model Size Comparison with LGBM}
Fig.~\ref{fig:downstream} compares AUPRC across all five pretrained models and LGBM for each dataset split. Some variations in performance were observed across the fine-tuned dataset splits, even for the same prediction task. For disease prediction (primary hypertension and chronic kidney disease), the 32M or 101M models were optimal, and for medication prediction (amlodipine and pregabalin), the 11M model was optimal. In each case, the task-optimal model consistently outperformed LGBM in AUPRC.

The pretraining time for the 11M model, which was optimal for the medication prediction tasks, was 53.9 h (Table~\ref{tab:scale}). Scaling up to the 101M model increased the pretraining time to 232.2 h; however, this additional computational cost did not yield further AUPRC improvements for these specific tasks.

\section{Discussion}
This study systematically evaluated the interplay among model size, task characteristics, and computational constraints for structured medical foundation models. The pretraining loss decreased consistently with model size, confirming a scaling trend consistent with the NLP domain. However, the downstream performance exhibited task-dependent saturation, with the optimal model size diverging across tasks.

For disease prediction, large models (32M--101M) achieved superior performance, suggesting that these tasks benefit from a greater model capacity. For medication prediction, performance was saturated in the mid-scale model (11M). As our timing analysis revealed, scaling from 11M to 101M increased the pretraining time from 53.9 to 232.2 h; however, this over four-fold increase in computational cost provided no additional predictive benefit. Medication initiation is largely governed by clinical practice guidelines, which may impose a stronger regularity on prediction targets. This suggests that for such regular tasks, the model capacity exceeds the information density of structured data at a relatively early point owing to the limited vocabulary and sparsity compared with free text.

These findings offer a critical counter-perspective to the prevailing ``bigger is always better'' paradigm observed in recent medical foundation models. For instance, prior studies exploring generative decoder-only models on longitudinal event sequences (e.g., Curiosity~\cite{curiosity_2025}) have reported continuous performance improvements with increasing model scale for autoregressive simulation tasks. In contrast, our results using encoder-only models highlight a distinct performance plateau for specific downstream discriminative tasks. This divergence suggests that while generative modeling of entire patient timelines may benefit from massive parameter spaces, targeted classification tasks---particularly highly regularized clinical interventions like medication initiation---reach a representation capacity ceiling at much smaller scales. Thus, our study indicates that architectural design (generative vs. discriminative) and specific task characteristics inherently dictate the optimal model capacity, challenging the uniform necessity of massive models in structured healthcare data.

These insights have important implications for the clinical deployment of constrained computational resources. Rather than uniformly adopting the largest model, biomedical practitioners should align both the model architecture and capacity with the nature of the target task. Tasks involving natural disease progression may benefit from larger models, whereas rule-based clinical interventions (e.g., medication, laboratory testing, or surgery) may saturate at smaller scales. For regular tasks such as medication prediction, fine-tuning from a mid-scale model can surpass the LGBM baseline while drastically reducing the pretraining time. These findings indicate that for structured medical foundation models, task-appropriate model size selection is the key to balancing prediction quality and computational cost.

Limitations include: (1)~reliance on claims/DPC data without laboratory values, vital signs, or clinical notes, which are available in other databases~\cite{MIMIC_IV_2023} and could further enrich foundation model representations~\cite{Moor_FMgeneralist_2023}; (2)~hospital-based data without guaranteed cross-facility longitudinal linkage~\cite{EHRcontinuity_2023}; (3)~evaluation limited to disease and medication prediction; different trends may emerge for other tasks such as readmission prediction; and (4)~single-dataset evaluation without external or prospective validation~\cite{MultiCenter_CLMBR_2024}.

\section{Conclusion}
We constructed structured medical foundation models at five scales (2.2M--101M parameters) and compared their downstream performances across disease onset and medication initiation predictions. The models were pretrained using a random sample (2.3 million patients, 32 hospitals) from a nationwide 519-hospital Japanese claims/DPC database and fine-tuned under limited-label conditions. Disease prediction benefited from large models (32M--101M), whereas medication prediction was saturated in the midscale model (11M), revealing that the benefit of model scaling is task-dependent. For each task, the best-performing model consistently outperformed the LGBM baseline. These results demonstrate that task-appropriate model size selection can prevent the excessive computational hours required to pretrain unnecessarily large models, offering practical guidance for deploying structured medical foundation models under computational constraints.

\section*{Acknowledgment}
This study is supported by JST Moonshot R\&D (JPMJMS2021), JST Research and Development Program for Next-Generation Edge AI Semiconductors (JPMJES2511), JSPS KAKENHI (JP25K00148, JP25H02626, JP26K14994), and a project (JPNP14004) commissioned by the New Energy and Industrial Technology Development Organization (NEDO). The authors also gratefully acknowledge the University of Tokyo for the use of the Miyabi system for primary analyses and the Institute of Science Tokyo for the use of the TSUBAME supercomputer for preliminary work.

\section*{Conflict of Interest}
M.~Nakamura, N.~Matsui, and A.~Hatakama are employees of Medical Data Vision Co., Ltd. (MDV); IQVIA Solutions Japan G.K.; and DeSC Healthcare, Inc., respectively. MDV provided the medical claims data used in this study. These co-authors contributed specialized expertise regarding the collection, structure, and extraction logic of medical claims data, and supported the construction of the analytical dataset. Although the study received support in the form of data provision, it was conducted independently by the authors. None of the affiliated companies had any role in the study design, data analysis, interpretation of the results, or the decision to publish. No other commercial or financial relationships are construed as potential conflicts of interest.

\bibliographystyle{IEEEtran}
\bibliography{refs}

\setupSupplement

\begin{center}
    {\Large \bfseries Supplementary Materials \par}
\end{center}
\vspace{1.5em}

\subsection{Summary Statistics of the Full Analysis Cohort ($N = 2{,}294{,}687$)}
Table~S1 summarizes the full analysis cohort ($N = 2{,}294{,}687$).
Fig.~S1a--S1d show distributions of sequence length, unique timestamps, 
observation duration, and age. Note that the sequence length reported here 
counts only the clinical event tokens (ICD-10 and YJ codes), excluding 
the initial control tokens ([CLS] and sex). Sequence length and unique timestamps 
exhibited heavy right tails, while age peaked at 27{,}000--30{,}000 
days (74--82 years), consistent with an older adult population.

\begin{table}[H]
\centering
\caption{Input-sequence statistics for the full analysis cohort ($N = 2{,}294{,}687$ patients)}
\label{tab:S1}
\begin{tabular}{lc}
\toprule
Statistic & Median [Interquartile Range] \\
\midrule
Sequence length (tokens)            & 22 [7--104]                   \\
Number of unique timestamps (days)  & 7 [2--25]                     \\
Observation duration (days)         & 249 [23--1{,}645]             \\
Age at first record (days)          & 26{,}335 [21{,}634--29{,}563] \\
\bottomrule
\end{tabular}
\end{table}

\begin{figure}[H]
\centering
\safeincludegraphics[width=0.95\columnwidth]{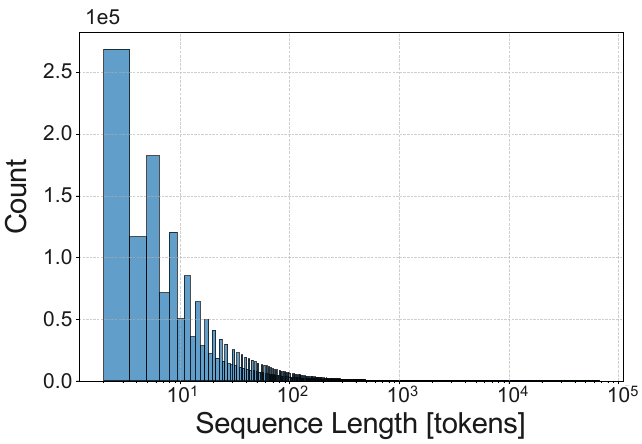}
\caption*{Fig.~S1a. Distribution of sequence length (tokens; x-axis on a logarithmic scale). The y-axis indicates patient count.}
\end{figure}

\begin{figure}[H]
\centering
\safeincludegraphics[width=0.95\columnwidth]{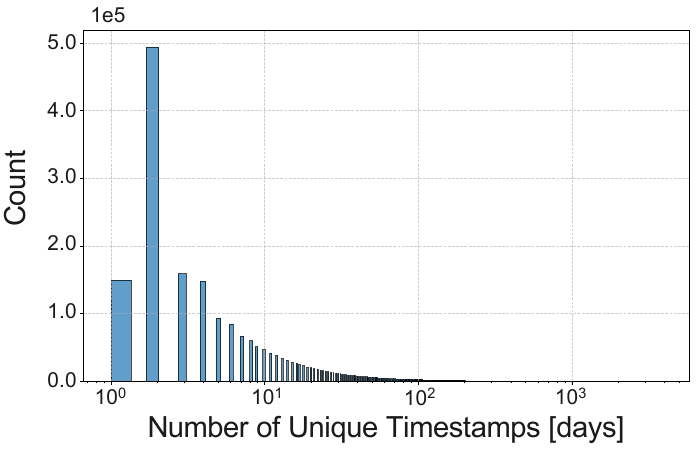}
\caption*{Fig.~S1b. Distribution of the number of unique timestamps (days; x-axis on a logarithmic scale). The y-axis indicates patient count.}
\end{figure}

\begin{figure}[H]
\centering
\safeincludegraphics[width=0.95\columnwidth]{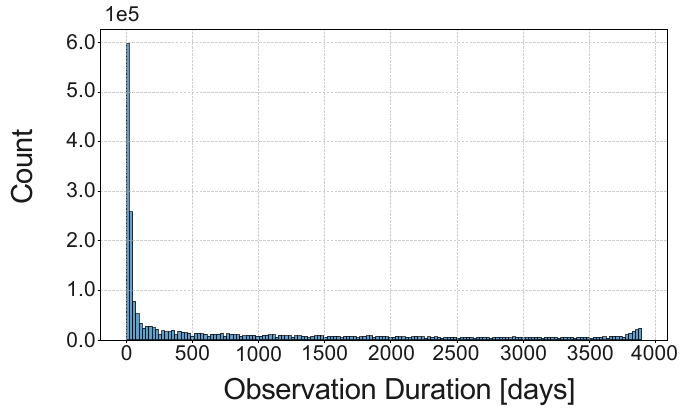}
\caption*{Fig.~S1c. Distribution of observation duration (days). The y-axis indicates patient count.}
\end{figure}

\begin{figure}[H]
\centering
\safeincludegraphics[width=0.95\columnwidth]{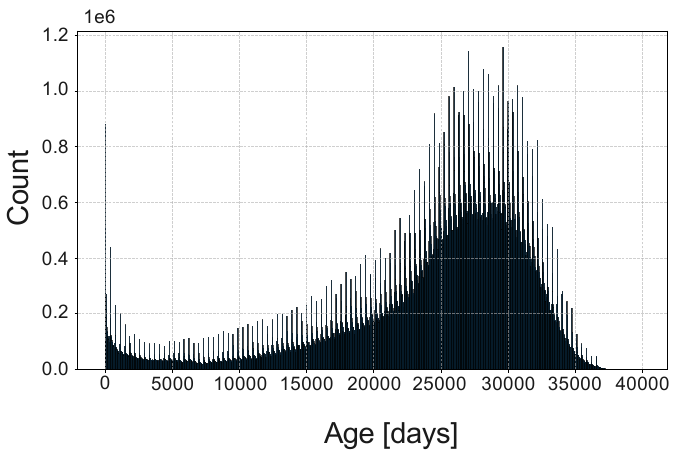}
\caption*{Fig.~S1d. Distribution of age in days. The y-axis indicates patient count.}
\end{figure}

\subsection{Pretrained Versus From-Scratch Models Across All Five Model Sizes (AUROC/AUPRC)}

Fig.~S2 extends the comparison between pretrained and from-scratch models, shown for the 11M model as a representative example in Fig.~4 of the main text, to all five model sizes (2.2M, 4.7M, 11M, 32M, and 101M). For each model size, test AUROC (left column) and test AUPRC (right column) are plotted against the number of labeled patients used for fine-tuning (100, 500, and 1{,}000). Pretrained models are shown in model-specific colors (2.2M = blue, 4.7M = orange, 11M = green, 32M = red, and 101M = purple), whereas from-scratch models are shown as gray-to-black dashed lines.

Across all model sizes, the performance gain associated with pretraining was more evident for AUPRC than for AUROC. In the fine-tuning datasets used in this study, the prevalence of positive labels ranged from 2.63\% to 14.9\%, indicating substantial class imbalance toward the negative class. The clearer improvement observed in AUPRC, a metric that more appropriately reflects performance under class imbalance, therefore provides practically important evidence for the value of pretraining in clinically relevant prediction tasks.

\begin{figure*}[p]
\centering
\safeincludegraphics[width=0.92\textwidth]{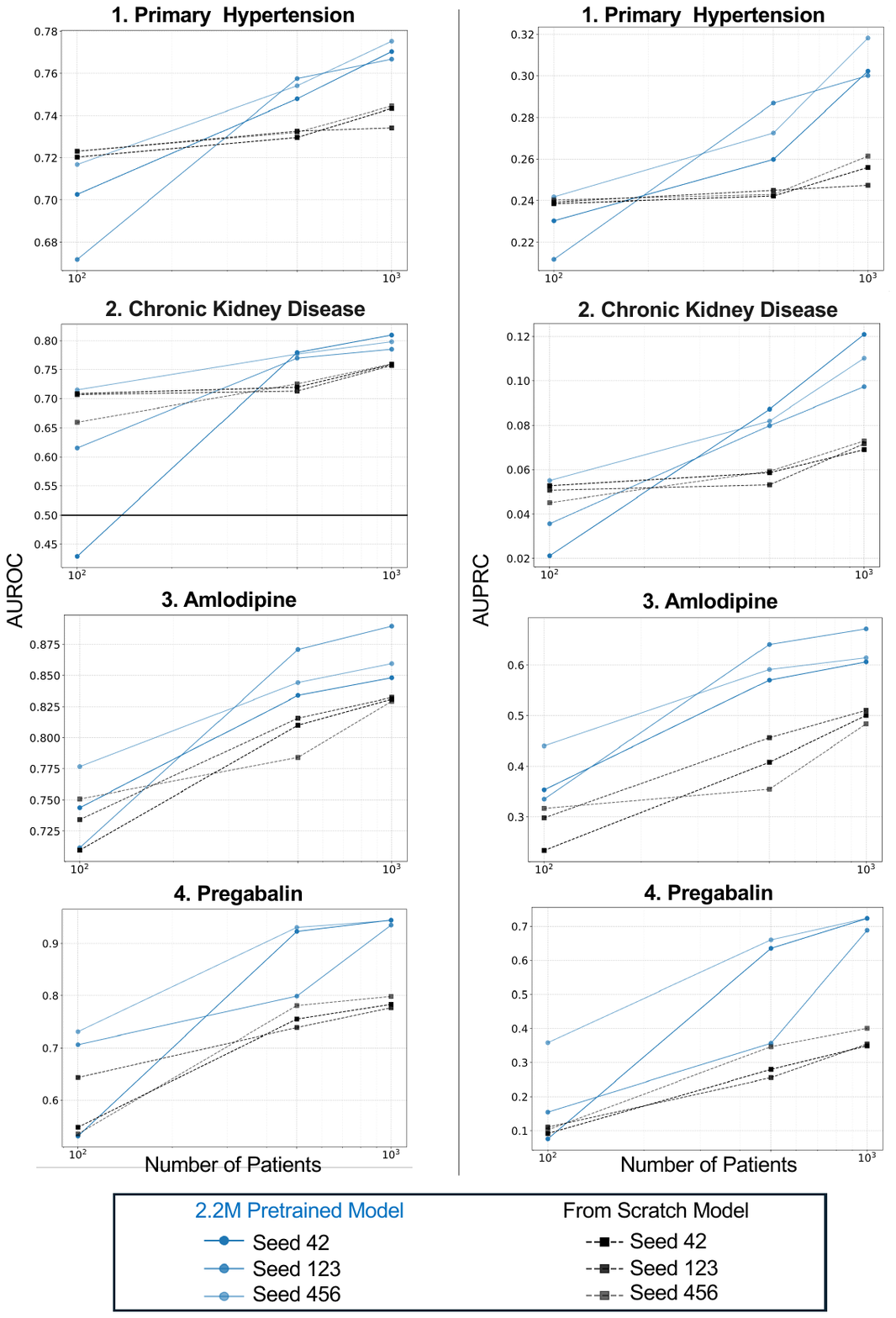}
\end{figure*}

\begin{figure*}[p]
\centering
\safeincludegraphics[width=0.92\textwidth]{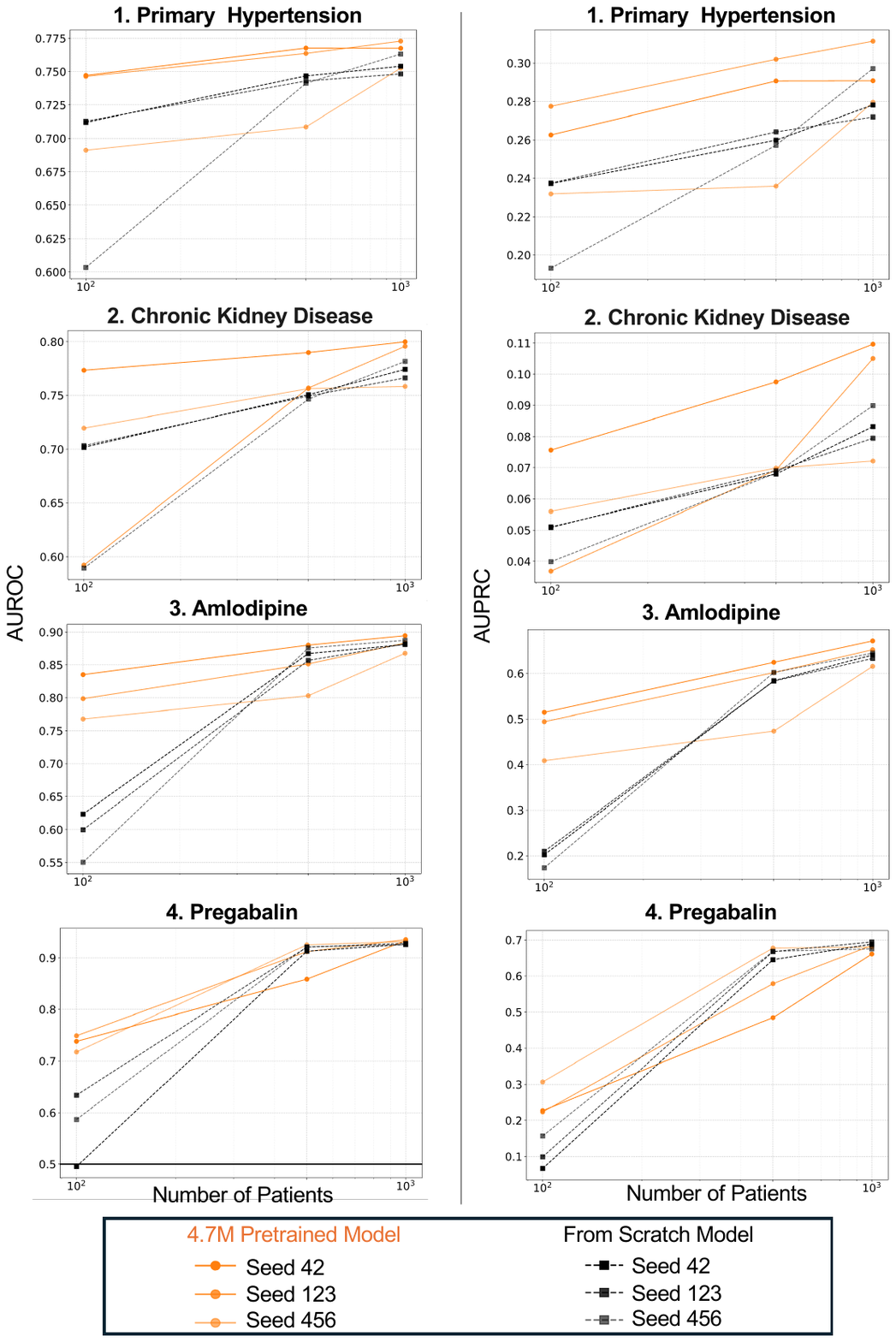}
\end{figure*}

\begin{figure*}[p]
\centering
\safeincludegraphics[width=0.92\textwidth]{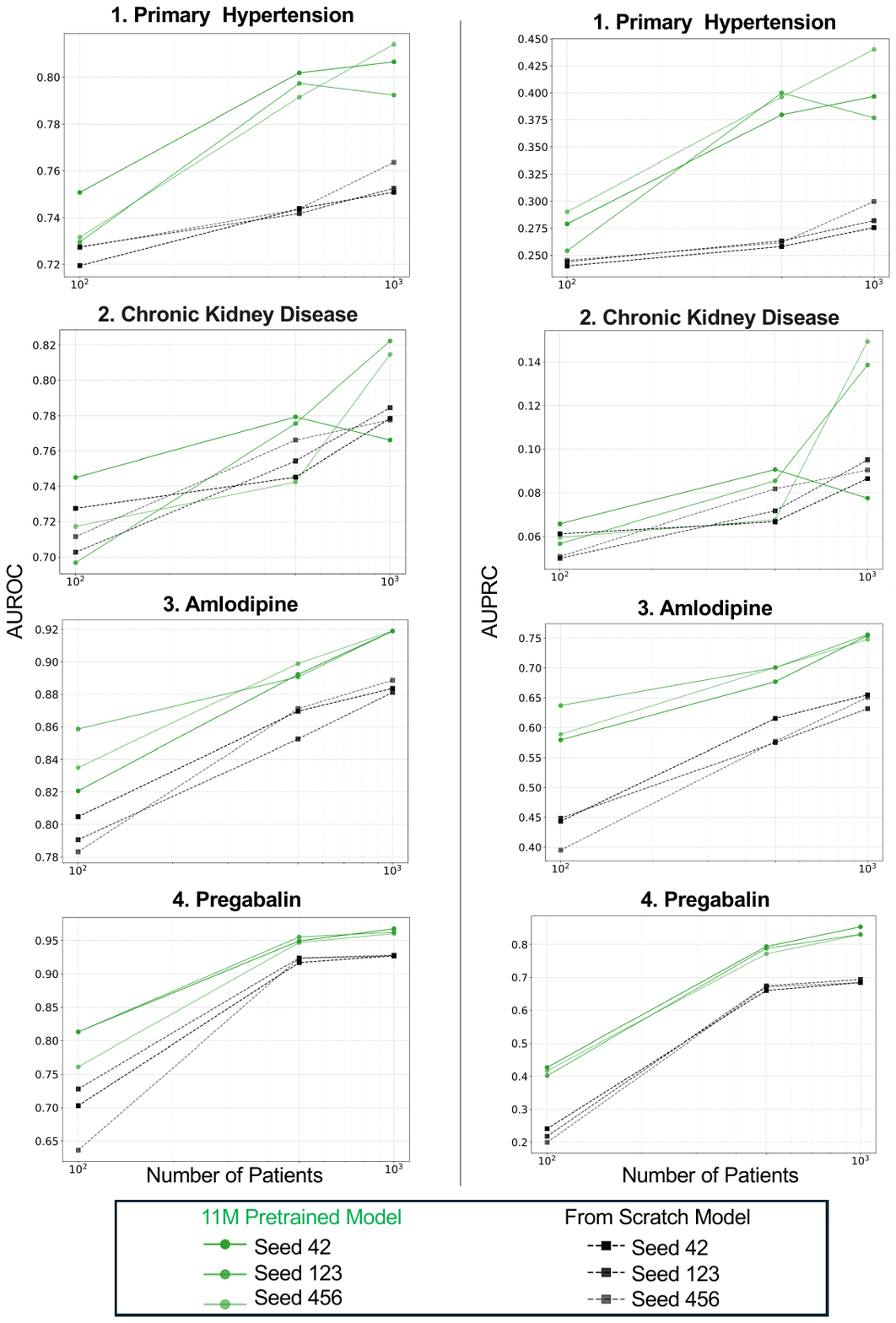}
\end{figure*}

\begin{figure*}[p]
\centering
\safeincludegraphics[width=0.92\textwidth]{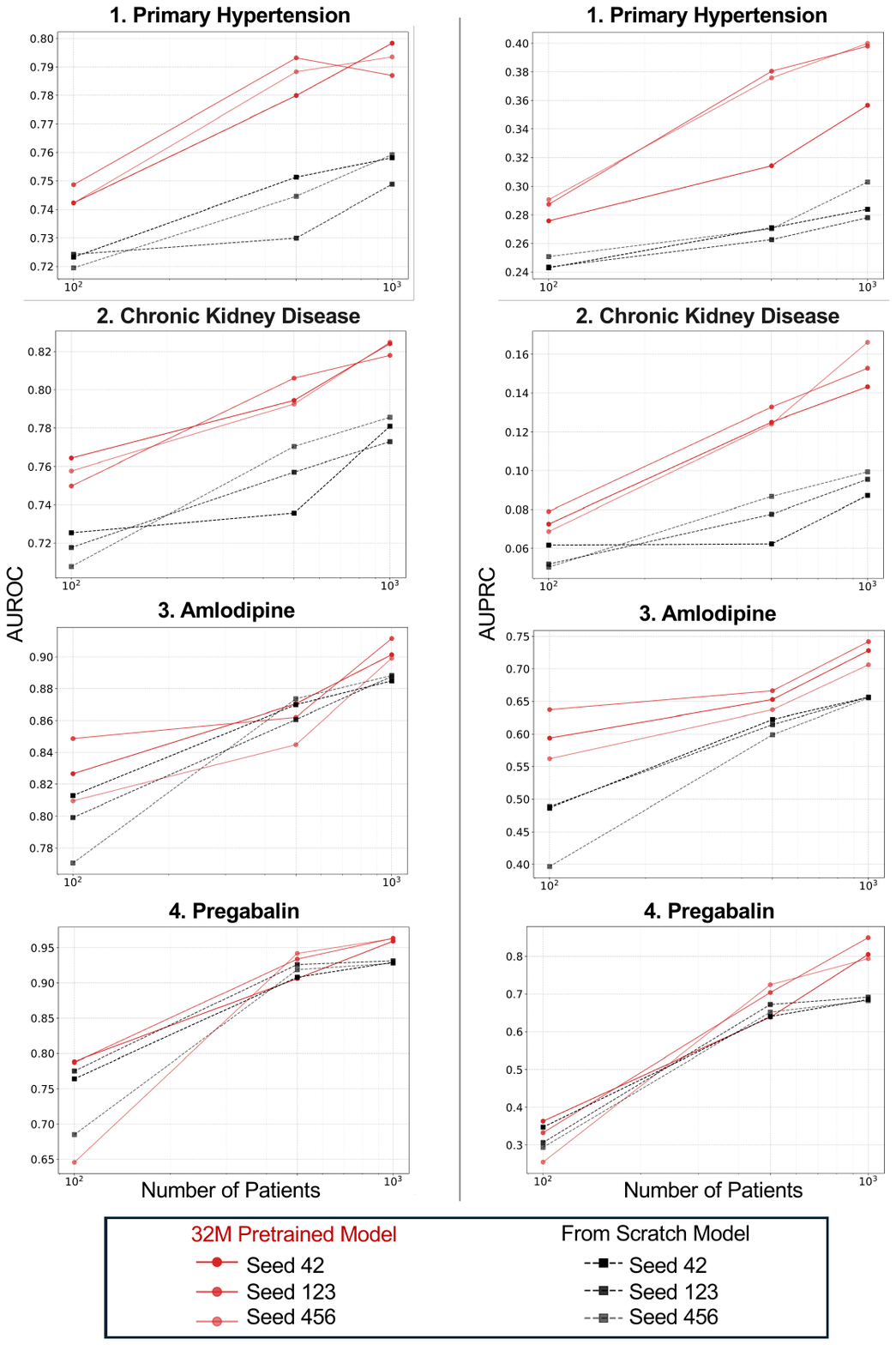}
\end{figure*}

\begin{figure*}[p]
\centering
\safeincludegraphics[width=0.92\textwidth]{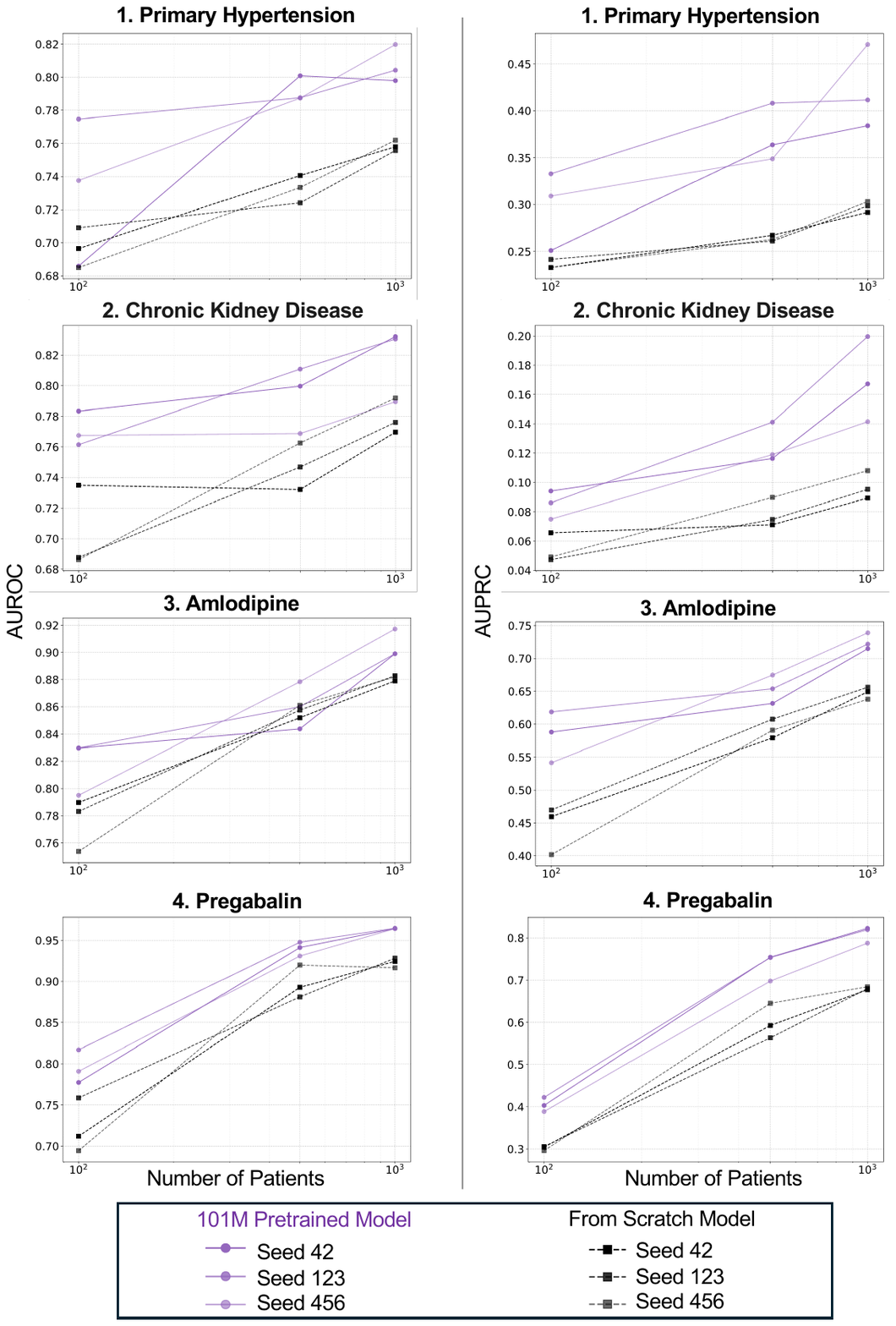}
\caption*{Fig.~S2. Comparison of test AUROC (left column) and test AUPRC (right column) between pretrained and from-scratch models across all five model sizes (2.2M = blue, 4.7M = orange, 11M = green, 32M = red, and 101M = purple). The x-axis indicates the number of patients with fine-tuning labels. Solid colored lines denote pretrained models, whereas gray-to-black dashed lines denote from-scratch models. Each line corresponds to one of three data splits (seed = 42, 123, or 456).}
\label{fig:S2}
\end{figure*}

\end{document}